\documentclass[10pt,twocolumn,letterpaper]{article}

\usepackage{iccv}
\usepackage{times}
\usepackage{epsfig}
\usepackage{graphicx}
\usepackage{amsmath}
\usepackage{amssymb}

\usepackage[pagebackref=true,breaklinks=true,letterpaper=true,colorlinks,bookmarks=false]{hyperref}

\iccvfinalcopy % *** Uncomment this line for the final submission

\def\iccvPaperID{6886} % *** Enter the ICCV Paper ID here

\ificcvfinal\fi

\begin{document}

%%%%%%%%% TITLE
\title{Self-Supervised Difference Detection\\
for Weakly-Supervised Semantic Segmentation}

\author{Wataru Shimoda ~and~ Keiji Yanai \medskip \\
Artificial Intelligence eXploration Research Center,
The University of Electro Communications, Tokyo\\
1-5-1 Chofugaoka, Chofu, Tokyo 182-8585 JAPAN\\
{\tt\small \{shimoda-k,yanai\}@mm.inf.uec.ac.jp}
% For a paper whose authors are all at the same institution,
% omit the following lines up until the closing ``}''.
% Additional authors and addresses can be added with ``\and'',
% just like the second author.
% To save space, use either the email address or home page, not both
%\and
%%Second Author\\
%%Institution2\\
%First line of institution2 address\\
%{\tt\small secondauthor@i2.org}
}

\maketitle
%\thispagestyle{empty}

%3/15 shimoda color blue

%%%%%%%%% ABSTRACT
\begin{abstract}
To minimize the annotation costs associated with the training of semantic segmentation models,
researchers have extensively investigated weakly-supervised segmentation approaches.
In the current weakly-supervised segmentation methods, the most widely adopted approach is based on visualization.
However, the visualization results are not generally equal to semantic segmentation.
Therefore, to perform accurate semantic segmentation under the weakly supervised condition, it is necessary to consider the mapping functions that convert the visualization results into semantic segmentation.
For such mapping functions, the conditional random field and iterative re-training using the outputs of a segmentation model are usually used.
However, these methods do not always guarantee improvements in accuracy; therefore, if we apply these mapping functions iteratively multiple times, eventually the accuracy will not improve or will decrease.

In this paper, to make the most of such mapping functions, we assume that the results of the mapping function include noise, and we improve the accuracy by removing noise.
To achieve our aim, we propose the self-supervised difference detection module, which estimates noise from the results of the mapping functions by predicting the difference between the segmentation masks before and after the mapping.
We verified the effectiveness of the proposed method by performing experiments on the PASCAL Visual Object Classes 2012 dataset, and we achieved 64.9\% in the val set and 65.5\% in the test set.
Both of the results become new state-of-the-art under the same setting
of weakly supervised semantic segmentation.
\end{abstract}

%%%%%%%%% BODY TEXT
\section{Introduction}
Semantic segmentation is a promising image recognition technology
that enables the detailed analysis of images for various practical applications. 
However, semantic segmentation methods require training data with pixel-level annotation, which is costly to create.
On the other hand, image-level annotation is much easier to obtain than pixel-level annotation.
In recent years, various weakly-supervised semantic segmentation (hereinafter WSS) methods that required only image-level annotation have been proposed to resolve the annotation problems. 
However, there is still a large performance gap between fully-supervised and weakly-supervised methods.

In weakly-supervised segmentation methods, visualization-based approaches~\cite{zei11, sim14, gap} have been widely adopted.
The visualization results highlight the regions that contributed to 
the classification, and we can roughly estimate the regions of the target objects by visualization.
Class Activation Map (CAM)~\cite{gap} is a standard method to visualize the classification results.
However, the visualization results do not always match actual segmentation results; therefore, it is usually necessary to consider the mapping from the visualization results to the semantic segmentation in weakly-supervised segmentation.
Conditional Random Field (CRF)~\cite{kra11}
is widely used as a mapping function.
CRF is a method for optimizing the probability distribution to be fitted to the edge of regions by using color and position information as features.
The iterative approach for the learning segmentation models proposed by Wei et al.~\cite{stc} is a versatile approach for improving weakly supervised segmentation results. 
In this method, we generate pseudo pixel-level labels under weakly supervised conditions, and we train a segmentation model with the pseudo labels. 
Subsequently, we generate pseudo pixel-level labels from the outputs of the trained segmentation model, and we re-train a new segmentation model using the generated pseudo labels. Wei et al.~\cite{stc} showed that repeating this process absorbed outliers and gradually improved the accuracy.
These methods can be regarded as mapping functions that bring inputs closer to the segmentation.
However, the mapping functions of these methods~\cite{kra11, stc} do not guarantee any improvement in the accuracy of the semantic segmentation; therefore, the mapping results contain noise.
In this paper, the mapping functions that make the above inputs close to the segmentation are treated as supervision containing noise, and we propose a robust learning method for such noise.

In this paper, we denote the information used as the inputs of the mapping functions as {\it knowledge}, and we consider the supervision containing the noise as {\it advice}. 
The supervision for fully supervised learning that allows one-to-one mapping is {\it teacher}.
We assume that the {\it advice} provides  supervision,  which  includes  some  correct  and incorrect information.
To make effective use of the information obtained from this {\it advice}, it is necessary to select useful information. 
In this paper, we regard the regions where opinions differ between {\it knowledge} and {\it advice} as {\it difference}.
Since {\it difference} in the two segmentation masks can be obtained by simple processing without annotation, it is a kind of self-supervised learning to train a model, which predicts {\it difference}.
Self-supervised learning is a pretext task as a form of indirect supervision.
For example, as notable works, colorization~\cite{dcolor} and predicting the patch ordering~\cite{porder} have been proposed.

\begin{figure*}[tb]
\begin{center}
\includegraphics[width=0.9\textwidth]{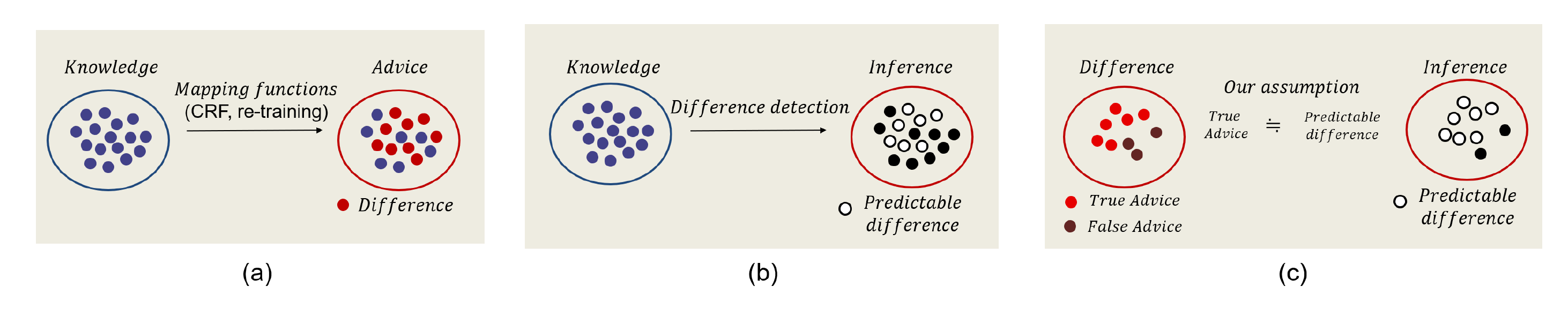}
\caption{The concept of the proposed approach.
(a) We denote the inputs of the mapping functions as {\it knowledge} and the outputs as {\it advice}.
(b) The proposed difference detection network~(DD-Net) estimates the {\it difference} between {\it knowledge} and {\it advice}.
(c) In {\it difference}, the {\it advice} is divided into true {\it advice} and false {\it advice}. We assume that if the amount of true {\it advice} is larger than the amount of false {\it advice}, that is, if a set of false {\it advice} are outliers, then the predictable {\it advice} has a strong correlation with the true {\it advice}.
} \label{mapping_fig}
\vskip -8mm~
\end{center}
\end{figure*}

Inferring {\it difference} in {\it knowledge} and {\it advice} from {\it knowledge} leads to predicting the advisor's {\it advice} in advance. 
In predicting {\it advice}, there are predictable {\it advice} and unpredictable {\it advice}.
Certain {\it advice} can be easily inferred because many similar samples are included during training.
Here, we assumed that {\tt advice} contains a sufficient number of good information, and predictable information can be considered to be useful information.
Based on this idea, we propose a method for selecting information by finding the true
information in {\tt advice} 
that can be predicted from the inference results of difference detection. 
Fig.\ref{mapping_fig} shows the concept of the proposed approach.

In this paper, we demonstrate that the proposed Self-Supervised Difference Detection~(SSDD) module can be used in both the seed generation stage and the training stage of fully supervised segmentation.
In the seed generation stage,
we refine the CRF results for pixel-level semantic affinity~(PSA)~\cite{psa} by using the SSDD module.
In the training stage, 
we introduce two SSDD modules inside the training loop of a fully supervised segmentation network.
In the experiments, we demonstrate the effectiveness of the SSDD modules in both stages.
In particular, the SSDD modules greatly boosted the performance of the WSS on the PASCAL visual object classes (VOC) 2012 dataset, and achieved new state-of-the-art.
To summarize it, our contributions are as follows:
\begin{itemize}
\item 
We propose an SSDD module, which estimates the noise of the mapping functions of the weakly supervised segmentation and select useful information. 
\item We show that the SSDD modules can be effectively applied to both the seed generation stage and
the training stage of a fully supervised segmentation model.
\item We obtained the best results on the PASCAL VOC 2012 dataset 
with 64.9\% mean IoU on the {\it val set} and 65.5\% on the {\it test set}.
\end{itemize}

\section{Related Works}
In this section, we review related research on CNN-based WSS methods by classifying them into several types.

\noindent
{\bf Visualization~~}
In the early works of CNN-based WSS, visualization-based methods were studied.
The pixels that contributed to the classification were correlated to the regions of the target objects; therefore, the visualization methods can be used as segmentation methods under weakly supervised settings.
Zeiler et al.~\cite{zei14} showed that the derivatives obtained 
by back-propagation from the CNN models trained for classification tasks
highlight the region of a target object in an image.
Simonyan et al.~\cite{sim14} used derivatives such as the GrabCut seeds and
extended the visualization method to the WSS method.
They also demonstrated that the regions of multi-class objects could also be captured 
by the difference in class-specific derivatives~\cite{backprop,dcsm}.
Oquab et al.\cite{oqu14} visualized the attention region by the forwarding process using activation and trained a classification model with large input images by using global max pooling.
After this approach, several derived methods employing global pooling were also
proposed~\cite{ped15,gap,sec}.
In particular, CAM~\cite{gap} has been widely adopted in recent weakly supervised segmentation methods.

\noindent
{\bf Region refinement for WSS results using CRF~~}
In general, the segmentation results based on fully convolutional neural network (FCN)~\cite{long15} tend to output ambiguous outlines.
CRF~\cite{kra11} can refine the ambiguous outlines using low-level features such as the pixel colors.
Chen et al.~\cite{papa15} and Pathak et al.~\cite{Pathak15} adopted CRF as a post-processing method for region refinement and demonstrated the effectiveness of the CRF for WSS.
Kolesnikov et al.~\cite{sec} proposed the use of CRF during the training of
a semantic segmentation model.
Ahn et al.~\cite{psa} proposed a method to learn pixel-level similarity from the CRF results,
and apply a random walk-based region refinement, 
which achieved the best results on the PASCAL VOC 2012 dataset.
CRF plays an important role to improve the accuracy of weakly supervised segmentation.
Furthermore, various researches employed the CRF for refining the coarse segmentation masks~\cite{dcsm,bfb,cbts,tphase,stc,erasing,seenet,cvpr18web}.
However, CRF does not guarantee any improvement in the mean intersection over union (IoU) score, 
and it often degrades the segmentation masks and the scores.
Therefore, we focus on preventing a segmentation mask from being degraded by applying CRF.
We estimate the confidence maps of both the initial mask and the mask after 
CRF post-processing, and we integrate both masks based on the estimated confidence maps.

\noindent
{\bf Training fully supervised segmentation model under weakly supervised setting~~}
Certain researchers trained a fully supervised semantic segmentation (hereinafter FSS) model 
under a weakly supervised setting.
First, Papandreou et al.~\cite{Pathak215} proposed MIL-FCN, which trained a fully supervised semantic segmentation model 
with a global max-pooling loss using only image-level labels.
Wei et al.~\cite{stc} proposed a novel approach to train an FSS model using pixel-level labels obtained by saliency maps~\cite{drfi}.
This method is simple, and the obtained results are impressive. 
Wei et al.~\cite{stc} also demonstrated that the outputs of the trained semantic segmentation model could be used as a new pixel-level annotation for re-training,
and the re-trained FSS model achieved better results than the original model.

\noindent
{\bf Generating pixel-level labels during training of an FSS model~~}
Constrained convolutional neural network (CCNN)~\cite{Pathak15} and EM-adopt \cite{papa15} generated pixel-level labels during training
using class labels and outputs of the segmentation model.
In both the studies similar constraints were made for 
generating pixel-level labels to obtain better results.
They set the ratios of the foreground and the background in an image and 
generated pixel-level labels within the ratio.
Wei et al.~\cite{erasing} proposed an online prohibitive segmentation learning~(PSL).
They generated pixel-level seed labels of training samples before the first training of an FSS model and re-generated pixel-level labels 
using the outputs of the segmentation model and the classification results.
The semantic segmentation model was trained by both the pixel-level labels, 
and they achieved good performance without costly manual pixel-level annotation.
We expected that the pixel-level seed labels would play the role of the constraint.
Huang et al.~\cite{dsrg} proposed deep seeded region growing~(DSRG), which is a method to expand the seed region during training.
Before training, the authors prepared pixel-level seed labels that had unlabeled regions for unconsidered pixels.
In this research, we proposed new constraints for generating pixel-level labels during the training of the FSS model. 
We trained an FSS model and the difference detection model in an end-to-end manner. 
Then, we interpolated a few pixel-level seed labels,
that had different regions in the newly generated pixel-level labels and these labels could also be predicted by the difference detection model.

\noindent
{\bf WSS methods using additional information~~}
A few recent weakly supervised approaches achieved high accuracy
by using additional annotations for image-level labels. 
Researchers have proposed the bounding box annotation for WSS
~\cite{papa15},
and they showed that the bounding box annotation substantially boosted performance.
As weaker additional annotation, 
point annotation and scribble annotation were also proposed~\cite{point}.
Saleh et al.~\cite{bfb} proposed an approach to check the generated
initial masks by minimal additional
supervision by human visions.
Motion segmentation of videos
as additional training information for weakly supervised segmentation has also been proposed~\cite{mcue,webvideo-seg}.
There are also reports that web images were helpful for improving the weakly supervised segmentation accuracy~\cite{ped15,stc,cvpr17web,cvpr18web}.
Recently, fully supervised saliency methods are being widely used for detecting the background regions, and certain researchers have reported that 
this approach could substantially boost performance~\cite{joon17cvpr, erasing, mdc, dsrg, seenet, mcof, dcsp}.
Region proposal methods trained with fully supervised foreground masks such as MCG~\cite{mcg} have also been used in \cite{ped15,afss}.
Hu et al.~\cite{salins} used instance-level saliency maps for WSS.
The concept of saliency can be used and helpful in various situation; however, the fully supervised saliency model was affected by its training data domain, which may cause negative effects on applications.
WSS methods without saliency maps are also beneficial.
In this paper, we do not use any additional information, and we use only PASCAL VOC images with image-level labels and CNN models pre-trained with ImageNet images and their image-level labels.

\section{Method\label{method}}
There was no supervision for the mapping functions of segmentation in the weakly supervised setting; therefore, it was necessary to consider a mapping for bringing the input close to the better segmentation results by using a method that incorporated human knowledge.
In this paper, we propose a method for selecting useful information from the results of the mapping functions by treating the results as supervision containing noise.
We define the inputs of the mapping functions as {\it knowledge}, and the mapped results as {\it advice}.
We predict the regions of {\it differences} between {\it knowledge} and {\it advice},
and we call this as the difference detection task. 
Using the inference results, we select the information of the {\it advice}.
\begin{figure}[tb]
\begin{center}
\includegraphics[width=0.48
\textwidth]{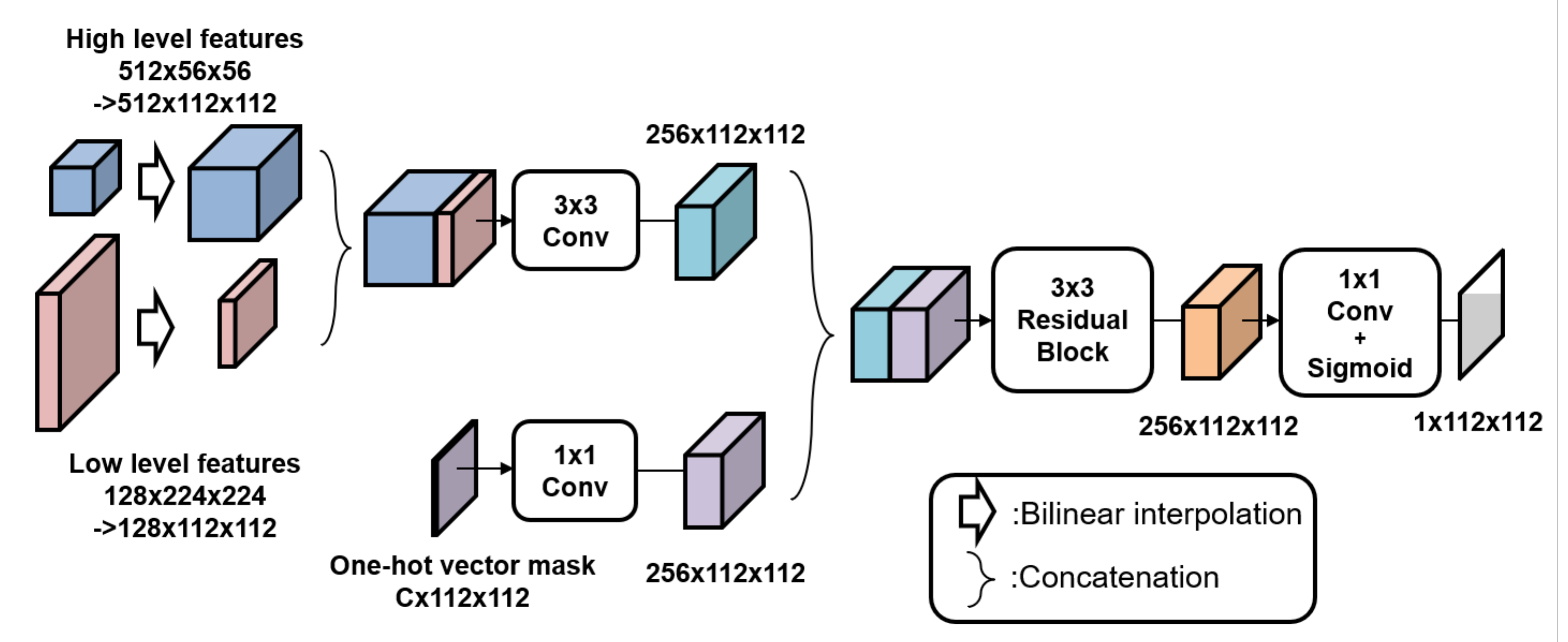}
\caption{Difference Detection Network~(DD-Net).} \label{fig_network}
\vskip -5mm~
\end{center}
\end{figure}

\subsection{Difference detection network}
In this section, we formulate the difference detection task.
In the proposed method, we predict the {\it difference} between {\it knowledge} and {\it advice}.
Here, we define the segmentation mask of {\it knowledge} as $ m^{K} $, the segmentation mask of {\it advice} as $ m^{A}$, and their {\it difference} as $ M^{K, A} \in {\mathbb R^{H \times W}} $.
\begin{equation}
 M^{K,A}_{u} =
\begin{cases}
  1&{\rm if}\hspace{1.5mm}(m^{K}_{u} = m^{A}_{u}) \\
  0&{\rm if}\hspace{1.5mm}(m^{K}_{u} \neq m^{A}_{u}) \\ 
\end{cases}
~~~,
\end{equation}
where $u \in \{1,2,..,n\}$ indicates a location of pixels, and $n$ is the number of pixels.
Next, we define a network of difference detection for deducing the {\it difference}.
We use feature maps extracted from a trained CNN to assist the difference detection.
In particular, we use high-level features $e^{h}(x;\theta_{e})$ and low-level features $e^{l}(x;\theta_{e})$
extracted from a backbone network, such as ResNet.
Here, $x$ is an input image, and $e$ is an embedding function parameterized by $\theta_{e}$.
As shown in Fig.\ref{fig_ssdd_module}, 
the confidence map of the input mask $d$ is generated by difference detection network~(DD-Net),
\footnotesize
${\rm DDnet}(e^{h}(x;\theta_{e}), e^{l}(x;\theta_{e}), \hat{m}; \theta_{d}), d \in {\mathbb R^{H \times W}}$~,
\normalsize
where 
$\hat{m}$ is a one-hot vector mask with the same number of channels to the target class number, $\theta_{d}$ is the parameter of the DD-Net, and $e(x)=(e^{l}(x),e^{h}(x))$.
The architecture of DD-Net is shown in Fig.\ref{fig_network}; it consists of three convolutional 
layers and one Residual block with three inputs and one output.
DD-Net takes either a raw mask or a processed mask as an input, and outputs the difference mask.
This network performs learning using the following losses:
\begin{equation}
\begin{split}
{\mathcal L}_{\mathit{diff}} = \frac{1}{|S|}\displaystyle{\sum_{u \in S}} (J(M^{K,A},d^{K},u;\theta_{d})\\
+ J(M^{K,A},d^{A}, u;\theta_{d})),
\end{split}
\end{equation}
where $S$ is a set of pixels of the input spaces, and $J()$ is assumed to be a function that returns a loss for the binary cross entropy.
\begin{displaymath}
 J(M,d,u)= M_{u}\log d_{u} +(1-M_{u})\log (1-d_{u}).
\end{displaymath}
Note that the parameters of the embedding function $\theta_{e}$ are independent of the optimization of $\theta_{d}$.
The training of DD-Net is self-supervised; therefore, neither special annotation nor additional data are needed.

\subsection{Self-supervised difference detection module}
\begin{figure}[tb]
\begin{center}
\includegraphics[width=0.48
\textwidth]{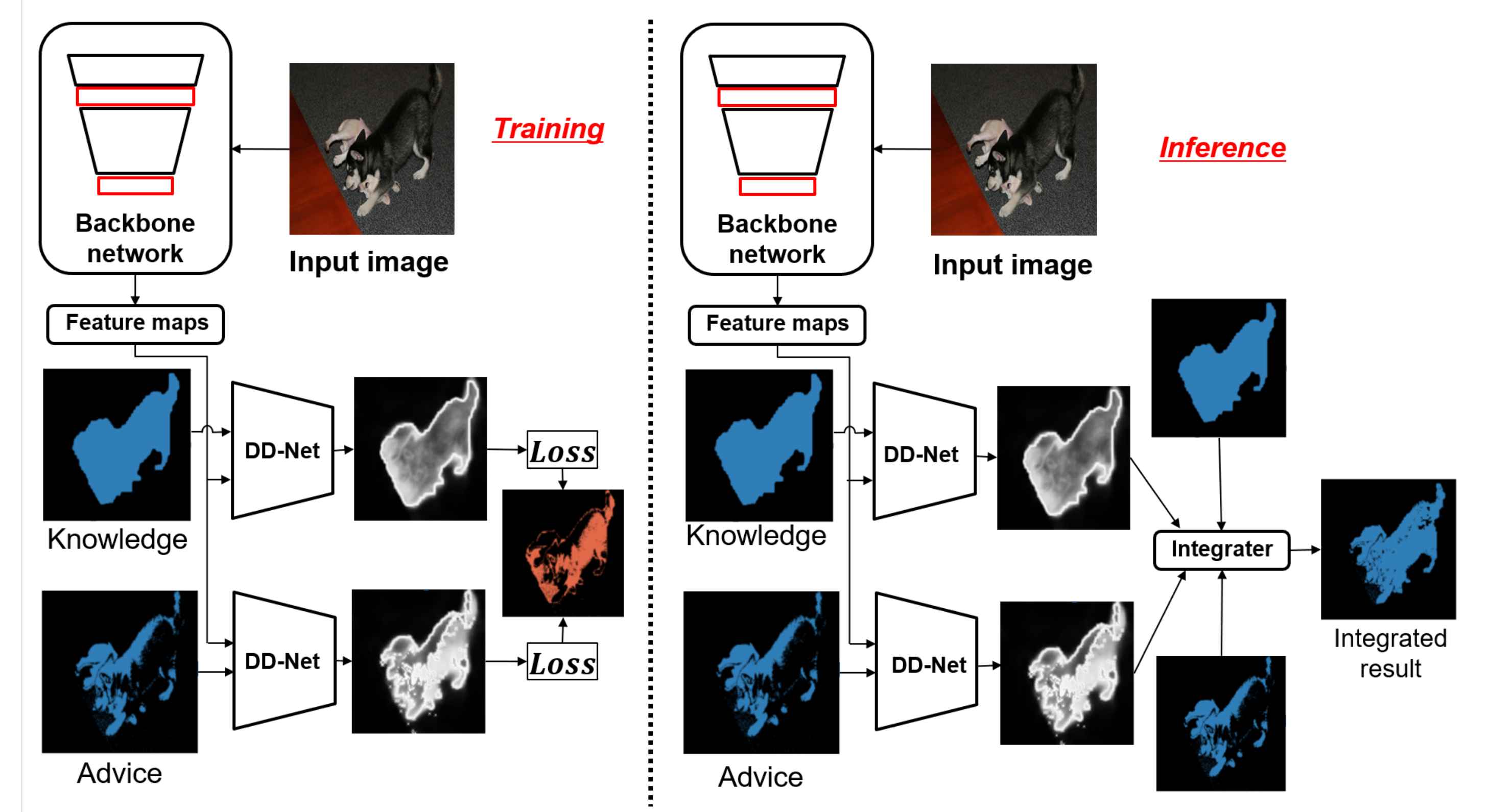}
\caption{Overview of the DD-Net. The figure on the left shows the training of the DD-Net, and the right figure shows the processing of the integration using the results of difference detection.} \label{fig_ssdd_module}
\vskip -5mm~
\end{center}
\end{figure}

In this section, we describe the details of the SSDD module shown in Fig.\ref{fig_ssdd_module},
which integrates two masks adaptively according to the confidence maps.
We denote a set of {\it advice} that are true in {\it difference} as $ S^{A, T}$, and a set of {\it advice} that are false as $S^{A, F}$.
The purpose of the method is to extract as many samples of $ S^{A, T} $ as possible from the entire set of {\it advice} $ S^{A} $. 
Let $ d^{K} $ be the inference results of {\it advice} from the given {\it knowledge}.
The inference results are the probability distributions from 0 to 1, and the values have variations.
The variations are caused by the difference in the difficulty of inference.
The presence of similar patterns during training can have a strong influence on the difference in the difficulty of inference. 
Here, if there are a sufficient number of {\it advice} that are true values rather than false values, that is, if $|S^{A,T}|>|S^{A,F}|$, the larger values indicate that their {\it advice} most likely belong to $S^{A,T}$.
However, for the values of $d^{K}$ at a boundary, it is not clear whether {\it advice}  belongs to $ S^{A, T} $ or not; this should probably be different from sample to sample.
Therefore, it is difficult to deduce a good {\it advice} directly from the size of the value of $ d^{K} $.
To alleviate the problem, we use the inference results about the state of {\it knowledge} for each {\it advice}.
Although {\it advice}s have large variations in their distribution, these variations are less than the variations in the distribution of {\it knowledge} in general.
Therefore, using {\it advice} to infer {\it knowledge} is assumed to be easier than using {\it knowledge} to {\it advice} inference.
In this paper, we consider the results of the inference of {\it knowledge} to {\it advice} for evaluating the difficulty of inference in each sample; we use the inferences for the thresholds for each sample.
Specifically, we calculate the confidence scores of {\it advice}  from the viewpoint of how close the values of $ d^{K} $ to $ d^{A} $.
The confidence score $w_{u} \in {\mathbb R}$ is defined by the following expression:
\begin{equation}
w_{u}=d^{K}_{u}-d^{A}_{u}+bias_{u}
\label{eq_w}
\end{equation}
Here, $bias$ is a hyper parameter for a threshold of the selection obtained by the difference detection, and it is also an enhanced value for the categories in the presence labels of the input image.
The refined masks $ m^{D} $ obtained from $ m^{K}$ and $m^{A} $ are defined by the following expression:
\begin{equation}
 m^{D}_{u} = 
\begin{cases}
  m^{A}_{u} & {\rm if}\hspace{1.5mm}(w_{u} \geq 0) \\
  m^{K}_{u} & {\rm if}\hspace{1.5mm}(w_{u} < 0) \\ 
\end{cases}
\end{equation}
We denote this processing flow for generating new segmentation mask as an SSDD module in the after notation.
\begin{equation}
 m^{D} = SSDD(e(x), m^{K},m^{A};\theta_{d})
\end{equation}

\section{Introducing SSDD modules into the processing flow of WSS \label{application}}
In this section, we explain how to use SSDD modules in the processing flow of WSS.
The proposed method can be adapted to various cases by applying inputs of the mapping function as {\it knowledge} and the results of the mapping function as {\it advice}.
The processing flow that we adopted in this paper consists of two stages:
the seed generation stage with static region refinement and 
the training stage of a segmentation model with dynamic region refinement.
In the first stage, we adapted the proposed method by applying the results of PSA as {\it knowledge} and its CRF results as 
{\it advice}~(Sec.\ref{static}).
In the second stage, we adapted the proposed method by applying the results of the first stage~(Sec.\ref{static}) as {\it knowledge}, and the outputs of the segmentation models trained by the masks were applied as {\it advice}~(Sec.\ref{dynamic}).

\subsection{Seed mask generation stage with static region refinement\label{static}}
\begin{figure}[tb]
\begin{center}
\includegraphics[width=0.4\textwidth]{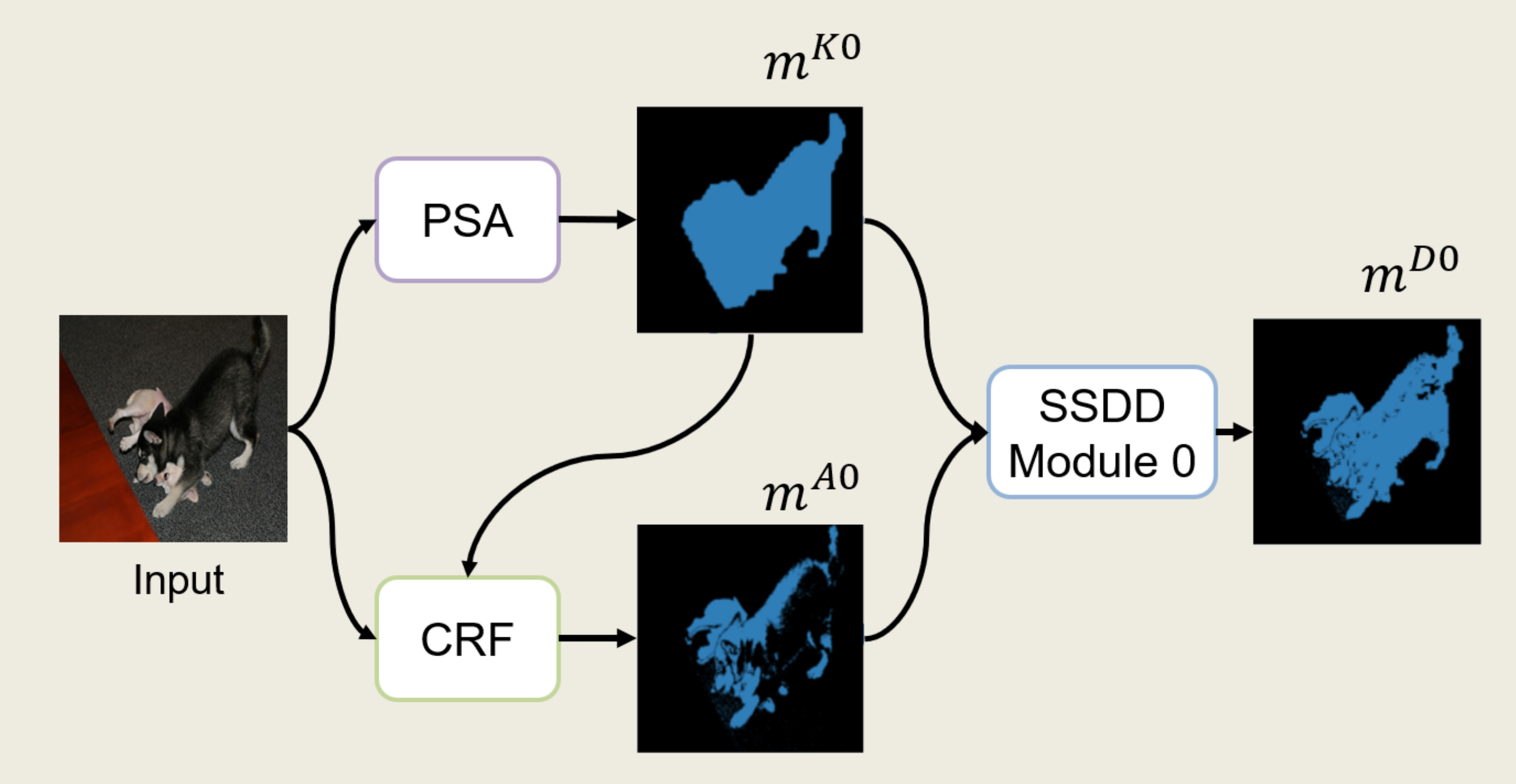}
\caption{Processing flow at the seed mask generation stage with static region refinement.} \label{fig_sssdd}
\vskip -5mm~
\end{center}
\end{figure}

PSA~\cite{psa} is a method to propagate label responses 
to nearby areas that belong to the same semantic entity.
Though PSA employs CRF for the refinement of the segmentation mask, CRF often fails to improve the segmentation masks; in fact, it degrades the masks.
In this section, we refine the outputs of CRF in PSA by using the proposed SSDD module.
We illustrate the processing flow of the first seed generation stage in Fig.\ref{fig_sssdd}.
Note that we omitted the input of the given image to an SSDD module for the sake of simplifying in the figure.

We denote an input image as $x$; the probability maps obtained by PSA are denoted as 
$p^{K0}=PSA(x;\theta_{psa})$, and its CRF results are denoted as $p^{A0}$.
We obtain the segmentation masks $(m^{K0},m^{A0})$ from the probability maps $(p^{K0},p^{A0})$ by taking the argument of the maximum of the presence labels including a background category.
We computed the loss of the DD-Net as follows:
\begin{equation}
\begin{split}
{\mathcal L}_{\mathit{diff0}} = \frac{1}{|S|}\displaystyle{\sum_{u \in S}} (J(M^{K0,A0},d^{K0},u;\theta_{d0})\\
+ J(M^{K0,A0},d^{A0}, u;\theta_{d0})),
\end{split}
\end{equation}

The proposed method is not effective when either of the segmentation masks or both of them do not have the correct labels.
These cases are not only meaningless for the proposed refinement approach, but they may also harm the training of the DD-Net. 
We define the bad training samples by simple processing based on the difference in the number of the class-specific pixels, and  we exclude them from the training.

In this work, we also train the embedding function by training a segmentation network with $m^{K0}$ to obtain good representation for the inputs of high-level features and low-level features:
\begin{equation}
{\mathcal L}^{base}={\mathcal L}^{seg}(x,m^{K0};\theta_{e0},\theta_{base}),
\end{equation}
\begin{equation}
{\mathcal L}^{seg}(x,m;\theta)=-\frac{1}{\displaystyle{\sum_{k\in K} |S_{k}^{m}|}} \displaystyle{\sum_{k\in K}}\displaystyle{\sum_{u \in |S_{k}^{m}|}}\log(h_{u}^{k}(\theta)),
\end{equation}
where $S_{k}^{m}$ is a set of locations that belong to the class $k$ on the mask $m$; $h_{u}^{k}$ is the conditional probability of observing any label $k$ at any location $u \in \{1, 2, . . . , n\}$; and ${\mathcal C}$ is a set of class labels.
$\theta_{e0}$ are parameters of embedding functions and $\theta_{base}$ are parameters for the segmentation branch. 
The training of $\theta_{e0}$ is independent of $\theta_{d0}$.
The final loss function for the static region refinement using the difference detection is as follows:
\begin{equation}
{\mathcal L}_{static}={\mathcal L}_{base}+{\mathcal L}_{\mathit{diff0}}.
\end{equation}

After training, we integrate the masks  $(m^{K0},m^{A0})$ and obtain the integrated masks $m^{D0}$ using the SSDD module with the trained parameter $\theta_{d0}$ as follows:
\begin{equation}
 m^{D0} = SSDD(e(x), m^{K0},m^{A0};\theta_{d0}).
\end{equation}

\subsection{Training stage of a fully supervised segmentation model with a dynamic region refinement \label{dynamic}}
When we train a fully supervised semantic segmentation model with pixel-level seed labels,
the accuracy of the seed labels directly effects the performance of the segmentation.
The performance gain is expected by replacing the seed labels to better the pixel-level labels during training.
In this study, we propose a novel approach to constrain the interpolation of the seed labels during the training of a segmentation model.
The idea of the constraint is to limit the interpolation of seed labels only to predictable regions of difference detection between newly generated pixel-level labels and seed labels.

\begin{figure}[tb]
\begin{center}
\includegraphics[width=0.5\textwidth]{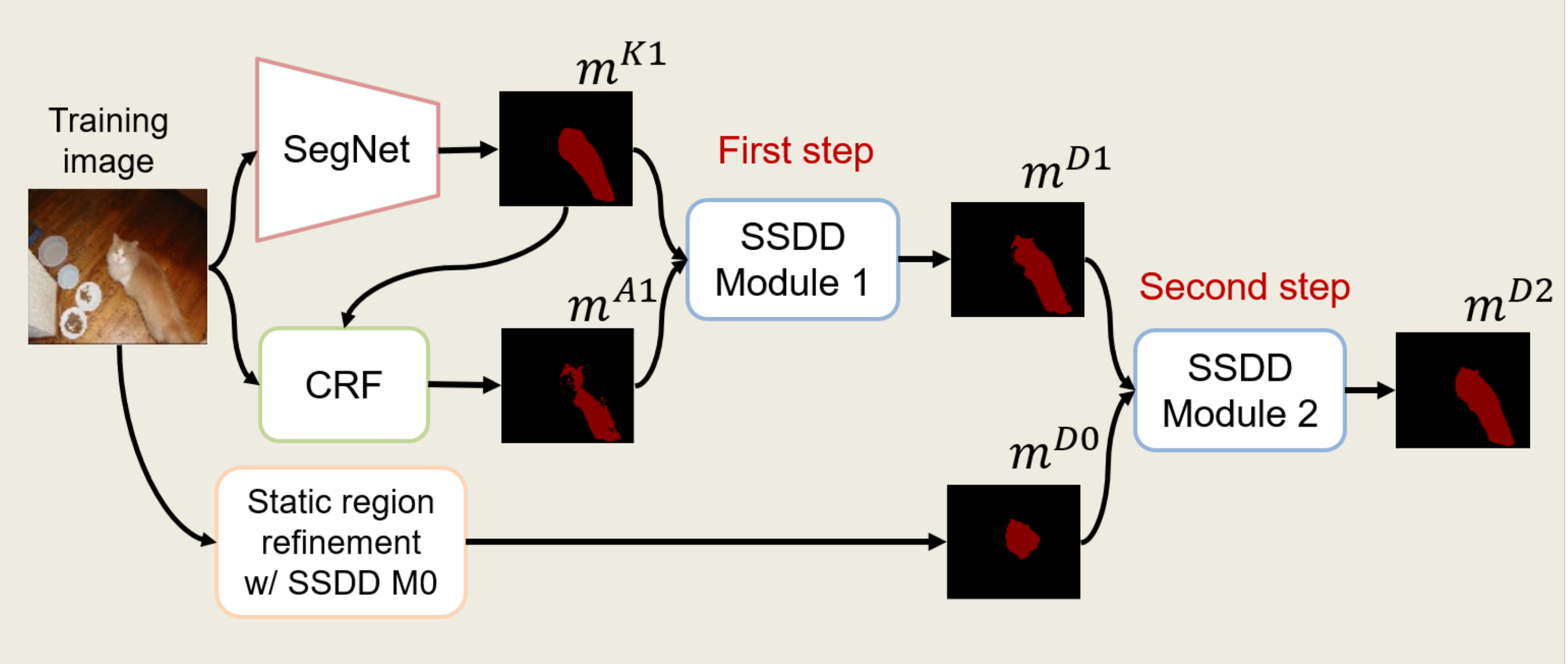}
\caption{
Illustration of the processing flow for the dynamic region refinement. 
(``SegNet'' does not represent any specific network but represents any kind of network for fully supervised semantic segmentation.)
} \label{fig_dssdd}
\vskip -5mm~
\end{center}
\end{figure}

In practice, we interpolate the pixel-level seed labels in two steps of each iteration as shown in Fig.\ref{fig_dssdd}.
Note that ``SegNet'' in the figure does not represent a specific segmentation network; it represents any fully supervised segmentation network. 
In the first step, for an input image $x$, we obtain the outputs of the segmentation model $p^{K1}=Seg(e(x);\theta_{main})$ and its CRF outputs $p^{A1}$. 
We obtain the segmentation masks $(m^{K1},m^{A1})$ from the probability maps $(p^{K1},p^{A1})$ by taking the argument of the maximum of the presence labels including a background category.
Then, we obtain the refined pixel-level labels $m^{D1}$ by applying the proposed refinement method as follows:
$m^{D1}=SSDD(e(x), m^{K1},m^{A1};\theta_{d1})$.
In the second step, we apply the 
proposed method to the seed labels $m^{D0}$ and to the mask $m^{D1}$ obtained in the first step.
The further refined mask $m^{D2}$ is obtained by
$m^{D2}=SSDD(e(x), m^{D0},m^{D1};\theta_{d2})$.
We generate the mask $m^{D2}$ in each iteration and train the segmentation model using the generated mask $m^{D2}$.
We train the semantic segmentation model with the generated mask $m^{D2}$ as follows:
\begin{equation}
{\mathcal L}_{main}= {\mathcal L}_{seg}(x, m^{D2};\theta_{e1}, \theta_{main}) ,
\label{seedloss}
\end{equation}

The loss of DD-Net for $ m ^ {A1} $ and $ m ^ {K1} $ is as follows:
\begin{equation}
\begin{split}
{\mathcal L}_{\mathit{diff1}} = \frac{1}{|S|}\displaystyle{\sum_{u \in S}} (J(M^{K1,A1},d^{K1},u;\theta_{d1})\\
+ J(M^{K1,A1},d^{A1}, u;\theta_{d1})),
\end{split}
\end{equation}
In the second stage,
we also exclude the bad samples (as done in Sec.{static}) based on the change ratio of pixels because the proposed method
is not effective if the input segmentation masks do not have correct regions.

We explain how to train the DD-Net for ($m^{D0}$,$m^{D1}$).
The masks ($ m ^ {K1}, m ^ {A1}, m ^ {D1} $) depend on the outputs of the segmentation model $ Seg (e (x), \theta_ {main}) $.
Therefore, if the learning of the segmentation model falls into a  local minimum, the masks will become meaningless; all the pixels become background pixels or single foreground pixels.
In this case, the inference results of the difference detection is also always constant, 
that is, ($ D ^ {K} = 1, d ^ {A} = 1, d ^ {A} = d ^ {K } $), and Eq.(\ref{eq_w}) becomes $w=bias$. 
To escape from this local minimum, we create a new branch of a segmentation model and use it for learning the difference detection between $ m ^ {D0} $ and $ m^{D1} $.
Assume that the mask $ m^{sub}$ was obtained from outputs of the branch of the new segmentation model $ p ^ {sub} = Seg (e(x); \theta_{sub})$.
In the training of difference detection, we trained the network to learn the differences among ($ m^{D0} $, $m^{sub}$) and ($m^{sub} $, $m^{D1}$) as follows:
\begin{equation}
\begin{split}
{\mathcal L}_{\mathit{diff2}} = \frac{1}{|S|}\displaystyle{\sum_{u \in S}} (J(M^{D0,sub},d^{D0},u;\theta_{d2})\\
+ J(M^{sub,D1},d^{D1}, u;\theta_{d2})),
\end{split}
\end{equation}
If $m^{sub}$ is the output, which is halfway between $m^{D0}$ and $m^{D1}$, the replacement of the training samples will let the segmentation model exit from the situation ($d^{K}=1, d^{A} = 1, d^{A} = d^{K}$), and the inference results of the difference detection will predict the regions that correlate with the {\it difference} between $m^{D0}$ and $m^{D1}$.
We train the parameters $ \theta_{sub} $ from the following loss to achieve the outputs that are halfway between $m^{D0}$ and $m^{D1}$.
\begin{equation}
{\mathcal L}_{sub}= \alpha{\mathcal L}_{seg}(x, m^{D0};\theta_{e1}, \theta_{sub}) + (1-\alpha){\mathcal L}_{seg}(x, m^{D1};\theta_{e1}, \theta_{sub}),
\label{seedloss2}
\end{equation}
where $\alpha$ is a hyper parameter of the mixing ratio of $m^{D0}$ and $m^{D1}$ .

The final loss function of the proposed dynamic region refinement method is calculated as follows:
\begin{equation}
{\mathcal L}_{dynamic}={\mathcal L}_{main} + {\mathcal L}_{sub}+ {\mathcal L}_{\mathit{diff1}}+ {\mathcal L}_{\mathit{diff2}} 
\end{equation}

\section{Experiments}
We evaluated the proposed methods using the PASCAL VOC 2012 data. 
The PASCAL VOC 2012 segmentation dataset has 1464 training images, 1449
validation images, and 1456 test images including 20 class pixel-level
labels and image-level labels. 
Similar to the methodology followed by~\cite{ped15,papa15,sec}, 
we used the augmented PASCAL VOC training data provided by \cite{Bha14} as well, wherein the training image number was 10,582.
For evaluation, we used an IoU metric, which is the official evaluation metric in the PASCAL VOC segmentation task.
For calculating the mean IoU on the val and test sets, we used the official evaluation server.
We compared the best performance of our method with the state-of-the-art methods on both the val and test sets.

\subsection{Implementation details}
Our experiments are heavily based on the previous research~\cite{psa}.
For the generating results of PSA results, we used implementations and trained parameters
provided by the authors that are publicly available.
We followed the methodology of~\cite{psa} and set hyperparameters
that gave the best performance.
For the CRF parameters, we used the default settings provided by \cite{kra11}.
For the semantic segmentation model, we used a ResNet-38 model, which had almost the same architecture as that in \cite{psa}.
The only difference was in the last upsampling rate; in the paper on PSA, the authors set the upsampling rate to 8, while we set the rate to 2 for reducing the computational cost of CRF.
The input image size was 448 for training, and the test images and the output feature map size before the upsampling was 56.
In the DD-Net, we used features obtained from the segmentation model before the last layer as the high-level features $e^{h}$ and the features obtained before the second pooling layer 
as the low level features $e^{l}$. 
These feature map sizes were adjusted to 112 by 112 using the simple linear interpolation approach.
We initialized the parameters of the segmentation models by using parameters trained with the PASCAL VOC images and their image-level labels with a pre-trained model using ImageNet, which was also provided in~\cite{psa}.
The codes provided by \cite{psa} did not include the training and test code for the segmentation models; therefore, we implemented our own codes.
In the original paper on PSA, though the authors optimized the segmentation models by Adam; however,
the performance was unstable in our re-implementation, and there were several unclear settings. 
Therefore, we used SGD for training the entire networks.
We set an initial learning rate to 1e-3 (1e-2 for initialization without the pre-trained model), and
we decreased learning rate with cosine LR ramp down~\cite{cosign_warm}.
For the static region refinement, we trained the network with batch sizes of 16 and 10 epochs.
For the dynamic region refinement, we trained the network with batch sizes of 8 and 30 epochs.
For the data augmentation and inference technique, we carefully followed the methodology used in~\cite{psa}.
We implemented the proposed method using PyTorch.
All the networks are trained using four NVIDIA Titan X PASCAL.
We will open the results of the proposed method and training codes\footnote{https://github.com/shimoda-uec/ssdd}.

\subsection{Analysis of static region refinement}
\begin{figure}[tb]
\begin{center}
\includegraphics[width=0.5\textwidth]{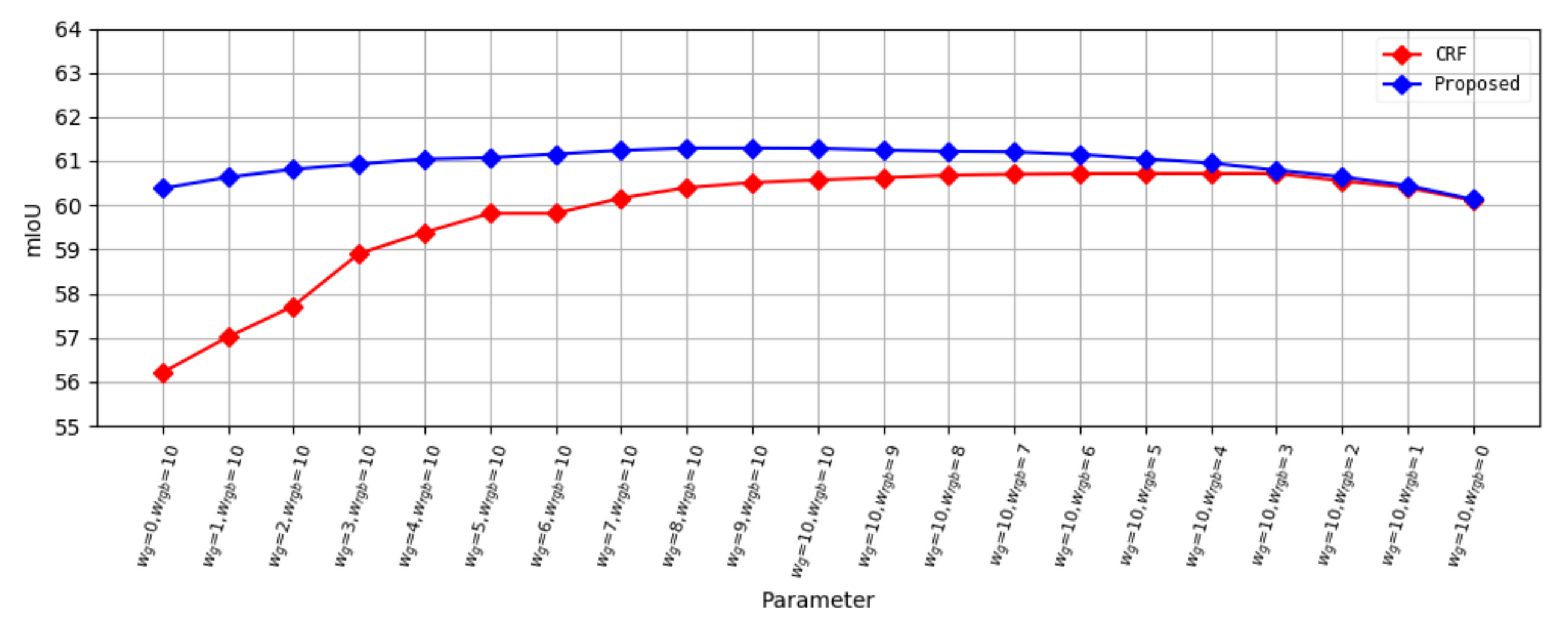}
\caption{mIoU of the seed masks of the training images with
different params values with only CRF and with SSDD and CRF.} \label{crf_cmp_fig}
\end{center}
\end{figure}

In the proposed method, we used fully connected CRF~\cite{kra11} with the same parameter settings as those for PSA~\cite{psa}, ($w_{g}=3$, $w_{rgb}=10$,$\theta_{\alpha}=80$, $\theta_{\beta}=13$, $\theta_{\gamma}=3$) in the following kernel potentials: $k(f_{i},f_{j})=w_{g}exp\left(-\frac{|p_{i}-p_{j}|}{2\theta_{\alpha}^{2}}-\frac{|I_{i}-I_{j}|}{2\theta_{\beta}^2}\right)+w_{rbg}exp\left(-\frac{|p_{i}-p{j}|^2}{2\theta_{\gamma}^{2}}\right)$.
To examine the relationship between the CRF params and results, we changed the values of ($w_{g}$, $w_{rgb}$) and evaluated the accuracy. 
Fig.\ref{crf_cmp_fig} shows a comparison of the proposed static region refinement with the PSA~\cite{psa} and its CRF results on the training set.
The weakening of $w_{rgb}$ decreases the difference only between the CRF and the SSDD+CRF results; therefore the effectiveness of the proposed method reduces. 
However, the proposed method always indicates a high accuracy. 
The optimal weights are different for each image, and it is expected to be difficult to search them for each image. 
We consider that the proposed method realized the improvement of CRF by correcting the partial failure of CRF.

\begin{figure}[tb]
  \begin{center}
\includegraphics[width=0.4\textwidth]{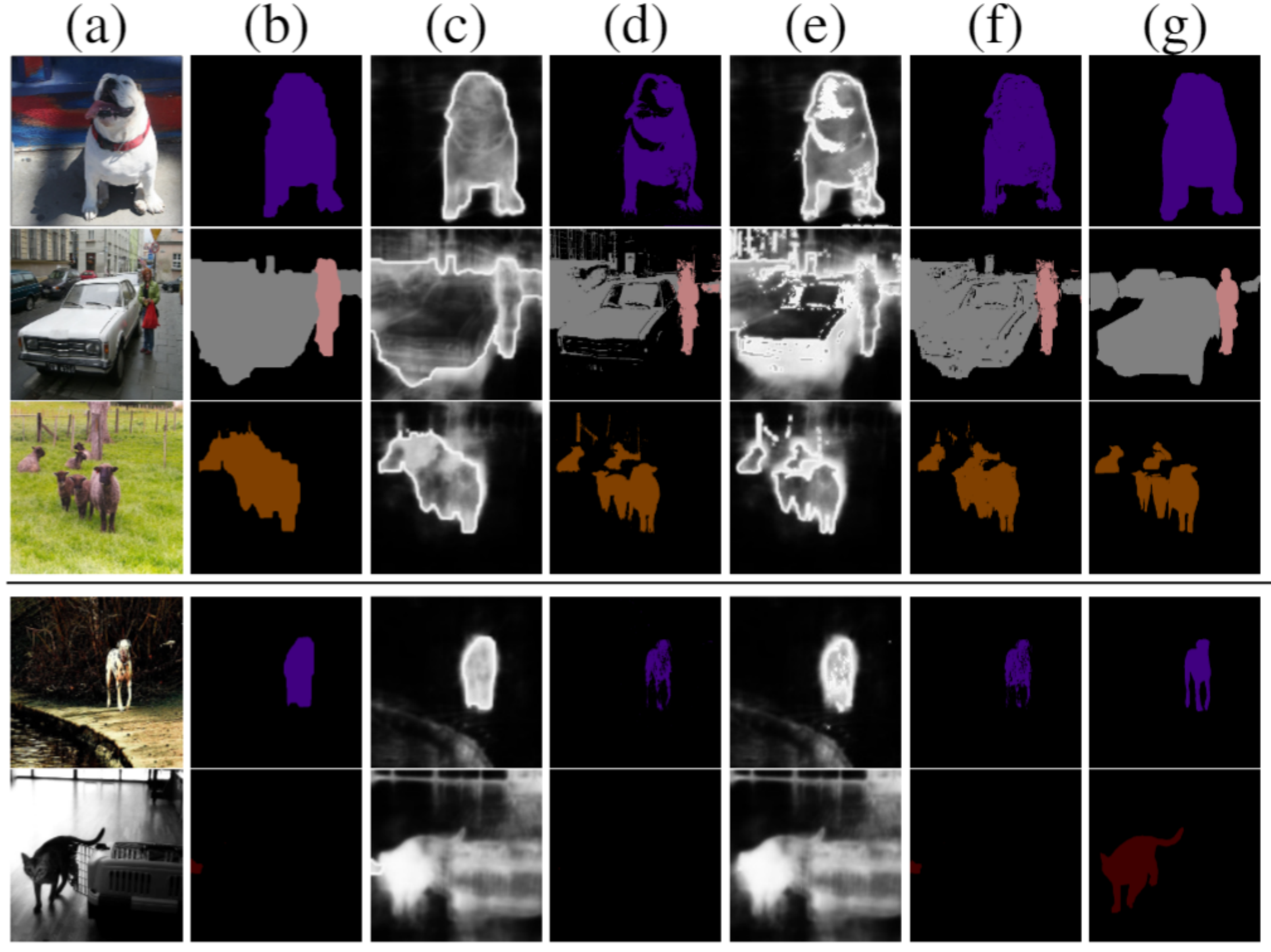}

\smallskip
\caption{Each row shows (a) input images, (b) raw PSA segmentation masks, (c) difference detection maps of (b), (d) CRF masks of (b), (e) difference detection maps of (d), (f) refined segmentation masks by the proposed method, and (g) ground truth masks.}
\label{exdp2}
\vskip -6mm~
\end{center}
\end{figure}

Fig.\ref{exdp2} shows the difference detection results and their refined segmentation masks.
In the fourth and fifth rows of Fig.\ref{exdp2}, we show the typical failure cases of the proposed method.
The regions of small objects tend to vanish in the CRF, and the DD-Net also learns such tendencies, which causes the failure of the proposed re-refinement method.
In the fifth row, both of the input segmentation masks fail to provide segmentation.
In such cases, the proposed method is also not effective.


\begin{table*}[tb]
\begin{center}
\caption{Results on PASCAL VOC 2012 {\it val set}. \label{tbl_dynamic}} 
\scalebox{1.0}[1.0]{
\scriptsize
\begin{tabular}[c]{|c|c*{20}{@{\hspace{0.07cm}}c@{\hspace{0.07cm}}}|c|}
\hline
Methods &\rotatebox{90}{Bg}& \rotatebox{90}{Aero}&
 \rotatebox{90}{Bike}& \rotatebox{90}{Bird}& \rotatebox{90}{Boat}&
 \rotatebox{90}{Bottle}& \rotatebox{90}{Bus}& \rotatebox{90}{Car}&
 \rotatebox{90}{Cat}& \rotatebox{90}{Chair}& \rotatebox{90}{Cow}&
 \rotatebox{90}{Table}& \rotatebox{90}{Dog}& \rotatebox{90}{Horse}&
 \rotatebox{90}{Motor}& \rotatebox{90}{Person}& \rotatebox{90}{Plant}&
 \rotatebox{90}{Sheep}& \rotatebox{90}{Sofa}& \rotatebox{90}{Train}&
 \rotatebox{90}{Tv}& \rotatebox{90}{\textbf{mIoU}} \\ \hline
\scriptsize{PSA \cite{psa}} & 88.2 & 68.2 & 30.6 & 81.1 & 49.6 & 61.0 & 77.8 & 66.1 & 75.1 & 29.0 & 66.0 & 40.2 & 80.4 & 62.0 & 70.4 & 73.7 & 42.5 & 70.7 & 42.6 & 68.1 & 51.6 & 61.7\\
\scriptsize{SSDD} & 89.0 & 62.5 & 28.9 & 83.7 & 52.9 & 59.5 & 77.6 & 73.7 & 87.0 & 34.0 & 83.7 & 47.6 & 84.1 & 77.0 & 73.9 & 69.6 & 29.8 & 84.0 & 43.2 & 68.0 & 53.4 & 64.9\\
\hline
\scriptsize{Gain} & {\it +0.8} &  {\it -5.7} &  {\it -1.7} &  {\it +2.6} &  {\it +3.3} &  {\it -1.5} &  {\it -0.2} &  {\it +7.6} &  \textcolor{red}{{\it +11.9}} &  {\it +5.0} &  \textcolor{red}{{\it +17.7}} &  {\it +7.4} &  {\it +3.7} &  \textcolor{red}{{\it +15.0}} &  {\it +3.5} &  {\it -4.1} &  \textcolor{blue}{{\it -12.7}} &  \textcolor{red}{{\it +13.3}} &  {\it +0.6} &  {\it -0.1} &  {\it +1.8} &  {\it +3.2} \\
\hline
\end{tabular}
}
\vskip -4mm~
\end{center}
\end{table*}

\subsection{Analysis of the whole proposed method}
We denote the dynamic region refinement as ``SSDD'' in all the tables.
The score of the SSDD is with the CRF with parameters ($w_{g}=3$, $w_{rgb}=10$) that are default values from the author's public implementation.
We also used the parameters for the CRF during training.

\noindent
{\bf Comparison with PSA~~}
Table~\ref{tbl_dynamic} shows the comparison of the dynamic region refinement method with the PSA.
We observe that the proposed method outperforms PSA by more than 3.2 point margins.
This clearly proves the effectiveness of the interpolation for the seed labels with the novel constraint by difference detection.
The accuracy is greatly improved as compared with the results of the static region refinement because of the increase in the number of good {\it advice} by end-to-end learning of the segmentation model, that is, $| S^{A1,T}| > | S^{A0, T}|  $.

In Table~\ref{tbl_dynamic}, we also show the gains between the proposed method and PSA for detailed analysis.
We obtain over 10\% gain on the cat, cow, horse, and sheep classes.
Interestingly, all the classes that gave the large gain belonged to the animal category.
However, in the potted plant, airplane, and person class objects, it was hard to improve the segmentation mask by using the proposed method.
In the proposed method, we considered the precondition that {\it advise}, which is a true value, was larger than the value that was not a true value($ | S ^ {A, T} | > | S ^ {A, F}| $).
When this precondition was satisfied, the accuracy of the classes improved.
If the precondition was not satisfied, the accuracy did not improve or the accuracy decreased.

\begin{figure}[tbh]
\begin{center}
\includegraphics[width=0.4\textwidth]{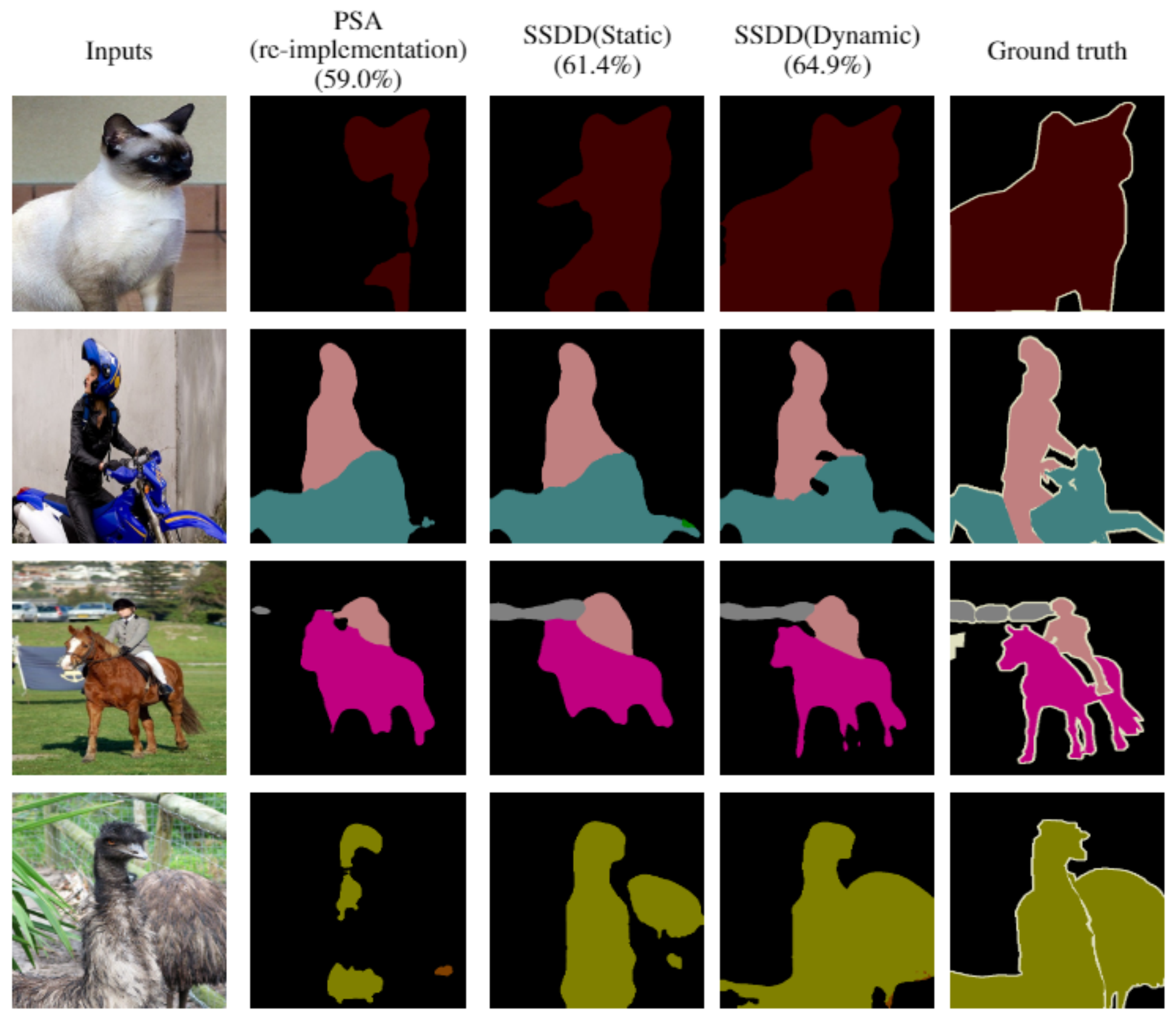}
\caption{Segmentation examples of results on PASCAL VOC 2012.} \label{exseg2}
\end{center}
\vskip -10mm~
\end{figure}

Fig.\ref{exseg2} shows the examples of the results of re-implementation of PSA, the static region refinement, and the dynamic region refinement.
Dynamic region refinement shows more accurate predictions on object location and
boundary.
The results of the static region refinement are outputs of a segmentation model re-trained with the masks in case of ($w_{g}=3$, $w_{rgb}=10$) in Fig.\ref{crf_cmp_fig}.
Note that we show the results of before the CRF for detailed comparisons.

\begin{table}[tb]
 \begin{center}
  \caption{Comparison with the WSS methods without additional supervision. \label{cmp_img_level}}
\scalebox{0.8}[0.8]{
  \begin{tabular}{l|c|cccc}
    Method  & Val & Test\\
      \hline
       FCN-MIL~\cite{Pathak215}\tiny{ICLR2015}    &  25.7 & 24.9 \\
       CCNN~\cite{Pathak15}\tiny{ICCV2015}    &  35.3 &35.6 \\
       EM-Adapt~\cite{papa15}\tiny{ICCV2015}      &  38.2 &39.6 \\
       DCSM~\cite{dcsm}\tiny{ECCV2016}     &  44.1 &45.1 \\
       BFBP~\cite{bfb}\tiny{ECCV2016}   &  46.6   &  48.0 \\
       SEC~\cite{sec}\tiny{ECCV2016}    &  50.7 &51.7 \\
       CBTS~\cite{cbts}\tiny{CVPR2017}    &  52.8 &53.7 \\
       TPL~\cite{tphase}\tiny{ICCV2017}   &  53.1 &53.8 \\
       MEFF~\cite{meff}\tiny{CVPR2018}   &  - &55.6 \\
       PSA~\cite{psa}\tiny{CVPR2018}      &  61.7 &63.7 \\
       IRN~\cite{irn}\tiny{CVPR2019} & 63.5 & 64.8 \\ \hline
       SSDD\tiny{ICCV2019}  & \bf 64.9 & \bf 65.5 \\
  \end{tabular}}
\smallskip
\caption{Comparison of the WSS methods with additional supervision. \label{cmp_add_info}}
\scalebox{0.75}[0.75]{
  \begin{tabular}{l|c|c|ccc}
    Method & Additional supervision & Val & Test\\
      \hline
       MIL-seg~\cite{ped15}\tiny{CVPR2015}  & Saliency mask + Imagenet images & 42.0  & 40.6\\
       MCNN~\cite{mcue}\tiny{ICCV2015}     & Web videos  &38.1   &  39.8 \\
       AFF~\cite{afss}\tiny{ECCV2016}    & Saliency mask & 54.3   &  55.5 \\
       STC~\cite{stc}\tiny{PAMI2017}    & Saliency mask + Web images & 49.8   &  51.2 \\
       Oh et al.~\cite{joon17cvpr}\tiny{CVPR2017}   & Saliency mask &  55.7   &  56.7 \\
       AE-PSL~\cite{erasing}\tiny{CVPR2017}    & Saliency mask & 55.0   &  55.7 \\
       Hong et al.~\cite{webvideo-seg}\tiny{CVPR2017}  & Web videos  &  58.1   &   58.7\\
       WebS-i2~\cite{cvpr17web}\tiny{CVPR2017}      & Web images & 53.4 &55.3 \\
       DCSP~\cite{dcsp}\tiny{BMVC2017}      & Saliency mask & 60.8 &61.9 \\
       GAIN~\cite{gain}\tiny{CVPR2018}     & Saliency mask & 55.3 &56.8 \\
       MDC~\cite{mdc}\tiny{CVPR2018}      & Saliency mask & 60.4 & 60.8 \\
       MCOF~\cite{mcof}\tiny{CVPR2018}    & Saliency mask & 60.3 &61.2 \\
       DSRG~\cite{dsrg}\tiny{CVPR2018}   & Saliency mask & 61.4 &63.2 \\
       Shen et al.~\cite{cvpr18web}\tiny{CVPR2018}   & Web images  &  63.0 & 63.9 \\
       SeeNet~\cite{seenet}\tiny{NIPS2018}    & Saliency mask & 63.1 &62.8 \\
       AISI~\cite{salins}\tiny{ECCV2018}      & Instance saliency mask & 63.6 &64.5 \\
       FickleNet~\cite{ficklenet}\tiny{CVPR2019} & Saliency mask & 64.9 & 65.3 \\ 
       DSRG+EP.~\cite{DSRGEP}\tiny{ICCV2019} & Saliency mask & 61.5 & 62.7 \\ 
       AttnBN.~\cite{AttnBN}\tiny{ICCV2019} & Saliency mask + Single-label images & 62.1 & 63.0 \\
       Zeng et al.~\cite{zeng_iccv2019}\tiny{ICCV2019} & Saliency mask & 63.3 & 64.3 \\ 
       OAA+.~\cite{OAA}\tiny{ICCV2019} & Saliency mask & 65.2 & 66.4 \\ 
       Lee et al.~\cite{lee_iccv2019}\tiny{ICCV2019} & Web videos & 66.5 & 67.4 \\ \hline
       SSDD\tiny{ICCV2019}    & - &  64.9 &  65.5 \\
  \end{tabular}
}
\vskip -8mm~
  \end{center}
\end{table}

\noindent
{\bf Comparison with the state-of-the-art methods~~}
Table~\ref{cmp_img_level} shows the results of the proposed method and the recent weakly supervised segmentation methods that do not use additional supervisions on the PASCAL VOC 2012 validation data and PASCAL VOC 2012 test data.
We observed that our method achieves the highest score as compared with all the existing methods, 
which use the same types of supervision~\cite{Pathak15,papa15,dcsm,bfb,sec,tphase,cbts,meff,psa}.
The proposed method outperforms the recent previous works on MEFF and TPL by large margins.
As discussed earlier, the proposed method also outperforms the current state-of-the-art methods~\cite{psa}.
This result clearly indicates the effectiveness of the proposed method.

Table~\ref{cmp_add_info} shows the comparison of the proposed method with 
a few weakly supervised segmentation methods that employ relatively cheap additional information.
Surprisingly, the proposed method also outperforms all the listed weakly supervised segmentation methods.
The proposed methods outperformed the following methods:
SeeNet~\cite{bfb}, DSRG~\cite{stc}, MDC~\cite{sec}, GAIN~\cite{gain}, and MCOF~\cite{mcof} that
employed fully supervised saliency methods.
In addition, the score of the proposed method was also better than the results of AISC~\cite{salins},
which used instance-level saliency map methods.
Note that AISC achieved 64.5\% on the val set and 65.6\% on the test set using an additional 24,000 ImageNet images for training.
The score of the proposed method was also higher than the score of Shen et al.~\cite{cvpr18web}, which used 76.7k web images for training.
It is not possible to have a completely fair comparison for them because of the difference of the network model, the augmentation technique, the number of iteration epochs, and so on.
However, the proposed method demonstrates comparable performance or better performance without any additional training information.

\section{Conclusions}
In this paper, we proposed a novel method to refine a segmentation mask from a pair of segmentation masks before and after the refinement process such as the CRF by using the proposed SSDD module.
We demonstrated that the proposed method could be used effectively in two stages: the static region refinement in the seed generation stage and the dynamic region refinement in the training stage.
In the first stage, we refined the CRF results of PSA~\cite{psa} by using the SSDD module.
In the second stage, we refined the generated semantic segmentation masks by using a fully supervised segmentation model and CRF during the training.
We demonstrated that three SSDD modules could greatly boost the performance of WSS and achieve the best results on the PASCAL VOC 2012 dataset over all the weakly supervised 
methods with and without additional supervision.

\medskip
\noindent 
{\bf Acknowledgements}~~
This work was supported by JSPS KAKENHI Grant
Number 17J10261, 15H05915, 17H01745, 17H06100 and 19H04929.

{\small
\bibliographystyle{ieee_fullname}
\bibliography{main}
}

\setcounter{table}{0}
\renewcommand{\thetable}{A-\arabic{table}}

\newpage
\onecolumn
\noindent
{\huge Supplementary Material for ``Self-Supervised Difference Detectionfor Weakly-Supervised Semantic Segmentation''}

\section*{A.1 Details of the simple decision}
In the proposed method, we select {\it advice} by inference results of difference detection.
The confidence score is calculated from the viewpoint of how close the value of $d^{K} $ to $d^{A} $. 
In the proposed method, if this difference is large enough, we ignore the {\it advice}.
Therefore, if the inferences of the difference detection are too easy, the values of $ d^{K} $ for {\it advice} that is not true become close to $d^{A} $, and the proposed method does not work effectively. 
In particular, if the inference results of the difference detection are ($d^{K} = 1, d^{A} = 1, d^{A} = d^{K}$), we cannot distinguish whether the {\it advice} belongs to the set of true values $|S^{A,T}|$ or the set of false values $|S^{A,F}|$ based on the results of the difference detection.
Therefore, we judge the typical failure examples of $ advice $ and excluded them from the training sample so that the differences between $d^{K}$ and $d^{A}$ were large in the inference of the bad {\it advice}.
To be concrete, when the number of differences in the pixels in each class of mask is obviously large, we assume that the {\it advice} has failed.
We define the bad training samples as the pair of the masks for the difference detection that satisfies the following equation:
\begin{equation}
\forall c \in {\cal C}, \frac{|S^{m^{A}}_{c}|}{|S^{m^{K}}_{c}|}<0.5,
\end{equation}
where ${\cal C}$ is a set of image-level label of the input image.
We decide the threshold 0.5 empirically.

\section*{A.2 Details of the bias in Eq.(\ref{eq_w})}
In Eq.(\ref{eq_w}), we use $bias$, which is a kind of hyperparameter.
In this section, we discuss this $bias$.
We define the $bias$ as follows:
\footnotesize
\begin{equation}
bias_{u} = 
\begin{cases}
b_{dd} \pm b_{class} & {\rm if}\hspace{1.5mm} m^{A}_{u}\hspace{1.0mm}or\hspace{1.0mm} m^{k}_{u}\hspace{1.0mm}belongs\hspace{1.0mm}to\hspace{1.0mm}\hat{{\cal C}}\hspace{1.0mm}\\
b_{dd} & {\rm if}\hspace{1.5mm}otherwise\\ 
\end{cases}~~~,
\label{eq_bias}
\end{equation}
\normalsize
where $\forall c \in \hat{{\cal C}}\hspace{1.5mm}$ satisfy$\hspace{1.5mm}\frac{|S^{m^{A}}_{c}|}{|S^{m^{K}}_{c}|}<0.5 \hspace{1.5mm}$and$\hspace{1.5mm}c\in {\cal C}$.
$b_{dd}$ is a bias for the difference between {\it knowledge} and {\it advice}, and $ b_{class} $ is a bias for the class category.
When the number of differences in the pixels in each class of mask is obviously large, it is assumed that the {\it advice} has failed, and to prioritize the label of that class over the results of the difference detection, we use the bias $b_{class}$.
We defined the values of $ b_ {dd}$ and $b_{class} $ by using the grid search.

\section*{A.3 Values of hyperparameters}

We explore good hyperparameters by a grid search and verify the effect of the hyperparameters.
We change the values of the hyperparameters and measure the mean IoU scores.
Table~\ref{hp} shows the hyper parameter values and the mean IoU scores.
The hyperparameters ($b_{dd}, b_{class}$) are used in Eq.(\ref{eq_bias}) as the bias values.
In $b_{dd}=0.4$, the mean IoU score becomes the maximum value.
We also set the bias $b_{class}$ for the missing categories.
We observe that the setting $b_{class}=1.0$ achieved a maximum mean IoU.
It is expected that the class biases for the missing categories help to the train for robustness.
In addition, we also verify the effect of hyperparameters for coefficients of losses in Eq.(\ref{seedloss}).
Though we had expected that the value of $\alpha$ would affect the performance, the hyper parameter was not critical for the change of the mean IoU.
The balanced setting, that is, $\alpha=0.5$ showed the best score.

\begin{table}[hb]
  \begin{center}
\caption{Experimental results with different parameters. \label{hp}}
\scalebox{0.8}[0.8]{
  \begin{tabular}{c|cccccccc}
     \hline
     $b_{dd}$ & 0.0 & 0.1 & 0.2 & 0.3 & \textbf{0.4} & 0.5  \\%l2
     mIoU & 62.2 & 63.9 & 64.6 & 64.2 & 64.9 & 62.7  \\%l2
     \hline
     $b_{class}$ & 0.0 & 0.5 & \textbf{1.0} & 1.5 & 2.0 \\ %l2
     mIoU & 64.3 & 63.0 & 64.9 & 64.5 & 63.7  \\%l2
     \hline
     $\alpha$ & 1.0 & 0.75 & \textbf{0.5} & 0.25&0.0  \\ %l2
     mIoU & 63.1 & 64.4 & 64.9 & 64.3  & 63.2  \\%l2
     \hline
  \end{tabular}
}
\vskip -8mm~
\end{center}
\end{table}

\newpage
\section*{A.4 Detailed comparison with existing works on the PASCAL VOC 2012 {\it val} and {\it test} sets}

\begin{table*}[htb]
\begin{center}
\caption{Results on PASCAL VOC 2012 {\it val set} without additional supervision. \label{table:val_a}} 
\scalebox{0.99}[0.99]{
\scriptsize
\begin{tabular}[c]{c|*{21}{@{\hspace{0.07cm}}c@{\hspace{0.07cm}}}|cc}
\hline
methods &\rotatebox{90}{bg}& \rotatebox{90}{aero}&
 \rotatebox{90}{bike}& \rotatebox{90}{bird}& \rotatebox{90}{boat}&
 \rotatebox{90}{bottle}& \rotatebox{90}{bus}& \rotatebox{90}{car}&
 \rotatebox{90}{cat}& \rotatebox{90}{chair}& \rotatebox{90}{cow}&
 \rotatebox{90}{table}& \rotatebox{90}{dog}& \rotatebox{90}{horse}&
 \rotatebox{90}{motor}& \rotatebox{90}{person}& \rotatebox{90}{plant}&
 \rotatebox{90}{sheep}& \rotatebox{90}{sofa}& \rotatebox{90}{train}&
 \rotatebox{90}{tv}& \rotatebox{90}{\textbf{mIoU}} \\ \hline
\scriptsize{MIL-FCN \cite{Pathak215}} & - & - & - & - & - & - & - & - & - & - & - & - & - & - & - & - & - & - & - & - & - & 24.9 \\
\scriptsize{CCNN \cite{Pathak15}} & 68.5 & 25.5 & 18.0 & 25.4 & 20.2 & 36.3 & 46.8 & 47.1 & 48.0 & 15.8 & 37.9 & 21.0 & 44.5 & 34.5 & 46.2 & 40.7 & 30.4 & 36.3 & 22.2 & 38.8 & 36.9 & 35.3 \\
\scriptsize{EM-Adapt \cite{papa15}} & - & - & - & - & - & - & - & - & - & - & - & - & - & - & - & - & - & - & - & - & - & 38.2 \\
\scriptsize{DCSM \cite{dcsm}}& 76.7 & 45.1 & 24.6 & 40.8 & 23.0 & 34.8 & 61.0 & 51.9 & 52.4 & 15.5 & 45.9 & 32.7 & 54.9 & 48.6 & 57.4 & 51.8 & 38.2 & 55.4 & 32.2 & 42.6 & 39.6 & 44.1 \\
\scriptsize{BFBP \cite{bfb}} &79.2 &60.1&20.4&50.7&41.2&46.3&62.6&49.2&62.3&13.3&49.7&38.1&58.4&49.0&57.0&48.2&27.8&55.1&29.6&54.6&26.6&46.6\\
\scriptsize{SEC \cite{sec}}&  82.4 & 62.9 & 26.4 & 61.6 & 27.6 & 38.1 & 66.6 & 62.7& 75.2 & 22.1 & 53.5 & 28.3 & 65.8 & 57.8 & 62.3 & 52.5 & 32.5 & 62.6 & 32.1 & 45.4 & 45.3&50.7\\
\scriptsize{CBTS~\cite{cbts}}    &  85.8 & 65.2 & 29.4 & 63.8 & 31.2 & 37.2 & 69.6 & 64.3 & 76.2 & 21.4 & 56.3 & 29.8 & 68.2 & 60.6 & 66.2 & 55.8 & 30.8 & 66.1 & 34.9 & 48.8 & 47.1 & 52.8 \\
\scriptsize{TPL~\cite{tphase}}   &  82.8 & 62.2 & 23.1 & 65.8 & 21.1 & 43.1 & 71.1 & 66.2 & 76.1 & 21.3 & 59.6 & 35.1 & 70.2 & 58.8 & 62.3 & 66.1 & 35.8 & 69.9 & 33.4 & 45.9 & 45.6 & 53.1 \\
\scriptsize{MEFF~\cite{meff}}   & - & - & - & - & - & - & - & - & - & - & - & - & - & - & - & - & - & - & - & - & - &- \\
\scriptsize{PSA~\cite{psa}} & 88.2 &68.2 &30.6 &81.1 &49.6 &61.0 &77.8 &66.1 &75.1 &29.0 &66.0 &40.2 &80.4& 62.0& 70.4 &73.7 &42.5 &70.7 &42.6 &68.1 &51.6 &61.7\\
\hline
\scriptsize{SSDD (ours)} & 89.0 & 62.5 & 28.9 & 83.7 & 52.9 & 59.5 & 77.6 & 73.7 & 87.0 & 34.0 & 83.7 & 47.6 & 84.1 & 77.0 & 73.9 & 69.6 & 29.8 & 84.0 & 43.2 & 68.0 & 53.4 & 64.9\\
\hline
\end{tabular}
}

\bigskip
\caption{Results on PASCAL VOC 2012 {\it test set} without additional supervision. \label{table:test}} 
\scalebox{0.99}[0.99]{
\scriptsize
\begin{tabular}[c]{c|*{21}{@{\hspace{0.07cm}}c@{\hspace{0.07cm}}}|cc}
\hline
methods &\rotatebox{90}{bg}& \rotatebox{90}{aero}&
 \rotatebox{90}{bike}& \rotatebox{90}{bird}& \rotatebox{90}{boat}&
 \rotatebox{90}{bottle}& \rotatebox{90}{bus}& \rotatebox{90}{car}&
 \rotatebox{90}{cat}& \rotatebox{90}{chair}& \rotatebox{90}{cow}&
 \rotatebox{90}{table}& \rotatebox{90}{dog}& \rotatebox{90}{horse}&
 \rotatebox{90}{motor}& \rotatebox{90}{person}& \rotatebox{90}{plant}&
 \rotatebox{90}{sheep}& \rotatebox{90}{sofa}& \rotatebox{90}{train}&
 \rotatebox{90}{tv}& \rotatebox{90}{\textbf{mIoU}} \\ \hline
\scriptsize{MIL-FCN \cite{Pathak215}} & - & - & - & - & - & - & - & - & - & - & - & - & - & - & - & - & - & - & - & - & - & 25.7 \\
\scriptsize{CCNN \cite{Pathak15}} & 68.5 & 25.5 & 18.0 & 25.4 & 20.2 & 36.3 & 46.8 & 47.1 & 48.0 & 15.8 & 37.9 & 21.0 & 44.5 & 34.5 & 46.2 & 40.7 & 30.4 & 36.3 & 22.2 & 38.8 & 36.9 & 35.3 \\
\scriptsize{EM-Adapt \cite{papa15}} & - & - & - & - & - & - & - & - & - & - & - & - & - & - & - & - & - & - & - & - & - & 39.6 \\
\scriptsize{DCSM \cite{dcsm}}& 78.1 & 43.8 & 26.3 & 49.8 & 19.5 & 40.3 & 61.6 & 53.9 & 52.7 & 13.7 & 47.3 & 34.8 & 50.3 & 48.9 & 69.0 & 49.7 & 38.4 & 57.1 & 34.0 & 38.0 & 40.0 & 45.1 \\
\scriptsize{BFBP \cite{bfb}} & 80.3 & 57.5 & 24.1 & 66.9 & 31.7 & 43.0 & 67.5 & 48.6 & 56.7 & 12.6 & 50.9 & 42.6 & 59.4 & 52.9 & 65.0 & 44.8 & 41.3 & 51.1 & 33.7 & 44.4 & 33.2 & 48.0 \\
\scriptsize{SEC \cite{sec}}&  83.5 & 56.4 & 28.5 & 64.1 & 23.6 & 46.5 & 70.6 & 58.5 & 71.3 & 23.2 & 54.0 & 28.0 & 68.1 & 62.1 & 70.0 & 55.0 & 38.4 & 58.0 & 39.9 & 38.4 & 48.3 & 51.7\\
\scriptsize{CBTS~\cite{cbts}}    &  85.7 & 58.8 & 30.5 & 67.6 & 24.7 & 44.7 & 74.8 & 61.8 & 73.7 & 22.9 & 57.4 & 27.5 & 71.3 & 64.8 & 72.4 & 57.3 & 37.0 & 60.4 & 42.8 & 42.2 & 50.6 & 53.7 \\
\scriptsize{TPL~\cite{tphase}}   &  83.4 & 62.2 & 26.4 & 71.8 & 18.2 & 49.5 & 66.5 & 63.8 & 73.4 & 19.0 & 56.6 & 35.7 & 69.3 & 61.3 & 71.7 & 69.2 & 39.1 & 66.3 & 44.8 & 35.9 & 45.5 & 53.8 \\
\scriptsize{MEFF~\cite{meff}}   &86.6 & 72.0 & 30.6 & 68.0 & 44.8 & 46.2 & 73.4 & 56.6 & 73.0 & 18.9 & 63.3 & 32.0 & 70.1 & 72.2 & 68.2 & 56.1 & 34.5 & 67.5 & 29.6 & 60.2 & 43.6 & 55.6 \\
\scriptsize{PSA~\cite{psa}} &89.1 & 70.6 & 31.6 & 77.2 & 42.2 & 68.9 & 79.1 & 66.5 & 74.9 & 29.6 & 68.7 & 56.1 & 82.1 & 64.8 & 78.6 & 73.5 & 50.8 & 70.7 & 47.7 & 63.9 & 51.1 & 63.7\\
\hline
\scriptsize{SSDD (ours)} & 89.5 & 71.8 & 31.4 & 79.3 & 47.3 & 64.2 & 79.9 & 74.6 & 84.9 & 30.8 & 73.5 & 58.2 & 82.7 & 73.4 & 76.4 & 69.9 & 37.4 & 80.5 & 54.5 & 65.7 & 50.3 & 65.5\\
\hline
\end{tabular}
}

\bigskip
\bigskip
\caption{Results on PASCAL VOC 2012 {\it val set} with additional supervision. \label{table:val}} 
\scalebox{0.99}[0.99]{
\scriptsize
\begin{tabular}[c]{c|c|*{21}{@{\hspace{0.07cm}}c@{\hspace{0.07cm}}}|cc}
\hline
methods &\rotatebox{90}{info type$\dagger$} &\rotatebox{90}{bg}& \rotatebox{90}{aero}&
 \rotatebox{90}{bike}& \rotatebox{90}{bird}& \rotatebox{90}{boat}&
 \rotatebox{90}{bottle}& \rotatebox{90}{bus}& \rotatebox{90}{car}&
 \rotatebox{90}{cat}& \rotatebox{90}{chair}& \rotatebox{90}{cow}&
 \rotatebox{90}{table}& \rotatebox{90}{dog}& \rotatebox{90}{horse}&
 \rotatebox{90}{motor}& \rotatebox{90}{person}& \rotatebox{90}{plant}&
 \rotatebox{90}{sheep}& \rotatebox{90}{sofa}& \rotatebox{90}{train}&
 \rotatebox{90}{tv}& \rotatebox{90}{\textbf{mIoU}} \\ \hline
\scriptsize{MIL-seg \cite{ped15}} & S & 79.6 & 50.2 & 21.6 & 40.6 & 34.9 & 40.5 & 45.9 & 51.5 & 60.6 & 12.6 & 51.2 & 11.6 & 56.8 & 52.9 & 44.8 & 42.7 & 31.2 & 55.4 & 21.5 & 38.8 & 36.9 & 42.0 \\ 
\scriptsize{MCNN~\cite{mcue}} & WV & 77.5 & 47.9 & 17.2 & 39.4 & 28.0 & 25.6 & 52.7 & 47.0 & 57.8 & 10.4 & 38.0 & 24.3 & 49.9 & 40.8 & 48.2 & 42.0 & 21.6 & 35.2 & 19.6 & 52.5 & 24.7 & 38.1 \\
\scriptsize{AFF~\cite{afss}} & S & - & - & - & - & - & - & - & - & - & - & - & - & - & - & - & - & - & - & - & - & - & 54.3 \\
\scriptsize{STC~\cite{stc}}   & S & 84.5 & 68.0 & 19.5 & 60.5 & 42.5 & 44.8 & 68.4 & 64.0 & 64.8 & 14.5 & 52.0 & 22.8 & 58.0 & 55.3 & 57.8 & 60.5 & 40.6 & 56.7 & 23.0 & 57.1 & 31.2 & 49.8 \\
\scriptsize{Oh et al.~\cite{joon17cvpr}}  & S  & - & - & - & - & - & - & - & - & - & - & - & - & - & - & - & - & - & - & - & - & - &  55.7 \\
\scriptsize{AE-PSL~\cite{erasing}}  & S & 83.4 & 71.1 & 30.5 & 72.9 & 41.6 & 55.9 & 63.1 & 60.2 & 74.0 & 18.0 & 66.5 & 32.4 & 71.7 & 56.3 & 64.8 & 52.4 & 37.4 & 69.1 & 31.4 & 58.9 & 43.9 & 55.0 \\
\scriptsize{Hong et al.~\cite{webvideo-seg}}  & WV & 87.0 & 69.3 & 32.2 & 70.2 & 31.2 & 58.4 & 73.6 & 68.5 & 76.5 & 26.8 & 63.8 & 29.1 & 73.5 & 69.5 & 66.5 & 70.4 & 46.8 & 72.1 & 27.3 & 57.4 & 50.2 & 58.1 \\
\scriptsize{WebS-i2~\cite{cvpr17web}} & WI & 84.3 & 65.3 & 27.4 & 65.4 & 53.9 & 46.3 & 70.1 & 69.8 & 79.4 & 13.8 & 61.1 & 17.4 & 73.8 & 58.1 & 57.8 & 56.2 & 35.7 & 66.5 & 22.0 & 50.1 & 46.2 & 53.4 \\
\scriptsize{DCSP~\cite{dcsp}}  & S & 88.9 & 77.7 & 31.3 & 73.2 & 59.8 & 71.0 & 79.2 & 74.5 & 80.0 & 15.1 & 73.3 & 10.2 & 76.1 & 72.2 & 69.1 & 72.1 & 39.9 & 73.9 & 14.6 & 70.3 & 53.1 & 60.8 \\
\scriptsize{GAIN~\cite{gain}}     & S & - & - & - & - & - & -  & - & - & - & - & - & - & - & - & - & - & - & - & - & - & -&56.8 \\
\scriptsize{MDC~\cite{mdc}}  & S& 89.5 & 85.6 & 34.6 & 75.8 & 61.9 & 65.8 & 67.1 & 73.3 & 80.2 & 15.1 & 69.9 & 8.1 & 75.0 & 68.4 & 70.9 & 71.5 & 32.6 & 74.9 & 24.8 & 73.2 & 50.8 & 60.4 \\
\scriptsize{MCOF~\cite{mcof}}  & S &  87.0 & 78.4 & 29.4 & 68.0 & 44.0 & 67.3 & 80.3 & 74.1 & 82.2 & 21.1 & 70.7 & 28.2 & 73.2 & 71.5 & 67.2 & 53.0 & 47.7 & 74.5 & 32.4 & 71.0 & 45.8 & 60.3 \\
\scriptsize{DSRG~\cite{dsrg}}   & S & - & - & - & - & - & - & - & - & - & - & - & - & - & - & - & - & - & - & - & - & -& 61.4 \\
\scriptsize{Shen et al.~\cite{cvpr18web}}  & WI  & 86.8 & 71.2 & 32.4 & 77.0 & 24.4 & 69.8 & 85.3 & 71.9 & 86.5 & 27.6 & 78.9 & 40.7 & 78.5 & 79.1 & 72.7 & 73.1 & 49.6 & 74.8 & 36.1 & 48.1 & 59.2 & 63.0 \\
\scriptsize{SeeNet~\cite{seenet}}   & S  & - & - & - & - & - & - & - & - & - & - & - & - & - & - & - & - & - & - & - & - & -& 63.1 \\
\scriptsize{AISI~\cite{salins}}      & IS & - & - & - & - & - & - & - & - & - & - & - & - & - & - & - & - & - & - & - & - & -  &64.5 \\ \hline
\scriptsize{SSDD (ours)}  & - & 89.0 & 62.5 & 28.9 & 83.7 & 52.9 & 59.5 & 77.6 & 73.7 & 87.0 & 34.0 & 83.7 & 47.6 & 84.1 & 77.0 & 73.9 & 69.6 & 29.8 & 84.0 & 43.2 & 68.0 & 53.4 & 64.9\\
\hline
\end{tabular}
}
\\
\footnotesize
($\dagger$ AS:Saliency mask, WV:web videos. WI Web images. IS Instance saliency mask.)
\normalsize
\end{center}
\end{table*}

\iffalse
\vskip 1.5cm~
\section*{B. Five hundred results on the validation set. }
The results are given in Table A-2.
In each row, 
an input image, a result by PSA~\cite{psa}, a result by SSDD with only first stage, a result by SSDD with both the first and second stage (full results), and ground truth are shown in the leftmost column. 
\fi

\end{document}